\documentclass[conference]{IEEEtran}

\usepackage{cite}
\usepackage{amsmath,amssymb,amsfonts}
\usepackage{algorithmic}
\usepackage{algorithm}
\usepackage{graphicx}
\usepackage{textcomp}
\usepackage{xcolor}
\usepackage{booktabs}
\usepackage{url}
\usepackage{pifont}
\usepackage{multirow}
\usepackage{enumitem}
\def\BibTeX{{\rm B\kern-.05em{\sc i\kern-.025em b}\kern-.08em
    T\kern-.1667em\lower.7ex\hbox{E}\kern-.125emX}}

\usepackage[hidelinks]{hyperref}
\definecolor{lb}{HTML}{e8f0fc}
\definecolor{pink}{HTML}{F5E3E3}
\usepackage{colortbl}
\usepackage{array}
\usepackage{flushend}

\begin{document}

\title{SMELL: Sparsity-aware Memory Efficient Federated Fine-tuning for Long-context LLMs\vspace{-0.05in}}

\author{\IEEEauthorblockN{Yin Xu\IEEEauthorrefmark{1}\IEEEauthorrefmark{2}, Yongbo Yang\IEEEauthorrefmark{1}\IEEEauthorrefmark{2}, He Sun\IEEEauthorrefmark{3}, Miao Sun\IEEEauthorrefmark{1}, Mingjun~Xiao\IEEEauthorrefmark{1}\IEEEauthorrefmark{2}}
\IEEEauthorblockA{\IEEEauthorrefmark{1}School of Computer Science and Technology, University of Science and Technology of China, Hefei, China}
\IEEEauthorblockA{\IEEEauthorrefmark{2}Suzhou Institute for Advanced Research, University of Science and Technology of China, Suzhou, China}
\IEEEauthorblockA{\IEEEauthorrefmark{3}Department of Data and Systems Engineering, The University of Hong Kong, Hong Kong, China}
Email: \{yinxu@, yyb118@mail.\}ustc.edu.cn, hesun@hku.hk, \{sm0602@mail., xiaomj@\}ustc.edu.cn}

\maketitle

\begin{abstract}
Federated fine-tuning enables Large Language Models (LLMs) to collaboratively adapt to distributed private data. However, long-context fine-tuning on resource-constrained devices faces substantial challenges arising from excessive activation memory, heterogeneous token-importance patterns across clients, and communication overhead. To address these challenges, we propose SMELL, a Sparsity-aware and Memory-Efficient federated fine-tuning framework for Long-context LLMs. SMELL integrates three complementary mechanisms: Local Sparse-Forward Adaptation (LSFA), Consensus-Anchored Sparsity Coordination (CASC), and Communication-Efficient Progressive Update Recovery (CPUR). These mechanisms jointly enable memory-efficient local adaptation, heterogeneity-aware sparsity coordination, and lightweight model-update transmission. Experiments on OPT-350M and LLaMA2-7B demonstrate that SMELL reduces peak memory consumption by up to 88.35\% compared with conventional fine-tuning and 13.25\% compared with forward-only LoRA fine-tuning at a 16K context length. Furthermore, SMELL improves fine-tuning performance under certain non-IID settings and supports compact update transmission with progressive reconstruction, demonstrating its potential for resource-efficient federated long-context LLM adaptation.
\end{abstract}

\begin{IEEEkeywords}
Long-context federated fine-tuning; large language models; zeroth-order optimization; token sparsity.
\end{IEEEkeywords}

\section{Introduction}

Large Language Models (LLMs) have demonstrated remarkable capabilities in a wide range of applications, including long-document understanding, multi-turn conversations, and domain-specific reasoning \cite{SurveyGPT, SurveyLLMEvaluation}. Supporting these applications often requires fine-tuning pretrained LLMs to long-context tasks using domain-specific data \cite{chen2023longlora,yang2023longqlora,ding2024longrope}. In many practical scenarios, however, such data are distributed across edge devices and cannot be readily centralized due to privacy and data governance concerns. Federated learning (FL) \cite{mcmahan2017communication, li2020federated, yang2019federated} provides a promising paradigm for collaborative model adaptation without requiring the collection of raw training data. LLM-driven FL paradigm is widely applied in fields such as medicine \cite{bai2026fedehr, nascimento2026federated, pan2025selective}, finance \cite{zhang2025incentivizing}, and robotics \cite{khalid2023cefhri}.

In practical FL scenarios, edge and mobile devices are typically resource-constrained, with limited memory, computational resources, and communication bandwidth. Consequently, enabling long-context LLM fine-tuning on such devices remains particularly challenging, as the memory required for intermediate activations grows rapidly with sequence length. Even parameter-efficient fine-tuning techniques such as LoRA \cite{hu2022iclr-lora}, which substantially reduce the number of trainable parameters, cannot fully alleviate this activation-memory bottleneck~\cite{wang2025jenga,tenison2026lars}. For instance, fine-tuning LLaMA2-7B \cite{Touvron2023Llama2} with a context length of 4K requires approximately 137 GB of memory, while LoRA-based fine-tuning still demands around 46 GB, far exceeding the memory budgets of devices. Moreover, the frequent transmission of model updates introduces substantial communication overhead, further straining devices with limited bandwidth. Thus, this paper focuses on the federated long-context LLM fine-tuning issue: how to enable resource-constrained devices to efficiently fine-tune long-context LLMs in federated scenarios, while simultaneously balancing model accuracy and communication overhead?

There are several challenges that need to be overcome in addressing the above issue under federated settings:

{\bf Challenge 1: Balancing memory efficiency and optimization reliability in long-context fine-tuning for resource-constrained devices.} Backpropagation-free (BP-free) optimization alleviates the need to retain backward activations, and token sparsification further reduces the memory footprint of attention computation \cite{wang2025jenga}. However, these techniques introduce different optimization trade-offs: Zeroth-Order Optimization (ZOO) methods such as FwdLLM \cite{FwdLLM} rely on stochastic gradient estimates \cite{Chen2017ZOO, malladi2023mezo}, whereas token sparsification changes the information available during forward computation. Effectively combining these complementary techniques while maintaining reliable optimization under stringent memory constraints remains a central challenge.

{\bf Challenge 2: Reconciling heterogeneous sparse token selection with globally shared model optimization.} Unlike centralized fine-tuning, FL distributes both data and training computation across devices with potentially non-IID distributions \cite{li2020federated, zhao2018federated}. Since token importance depends on the local input, independently applying contextual sparsification may lead to different token selection patterns across devices. Such differences may introduce additional discrepancies in the information used to construct local model updates, potentially affecting global optimization. Meanwhile, enforcing identical sparse masks can suppress information that is particularly important to individual devices. A key challenge is therefore to coordinate globally useful sparse selections while preserving flexibility for device-specific token importance, without requiring raw data centralization.

{\bf Challenge 3: Preserving aggregated update quality under constrained communication.} ZOO-based methods typically require multiple stochastic gradient estimates to reduce estimation variance \cite{FwdLLM, Chen2017ZOO}. In federated settings, the resulting high-dimensional model updates must also be communicated to a central server. Although gradient compression \cite{rothchild2020fetchsgdcommunicationefficientfederatedlearning} can reduce transmission payloads, it introduces reconstruction errors in addition to the stochastic estimation errors inherent in ZOO-based methods. Thus, a key challenge is how to leverage repeated gradient estimation to design communication-efficient update representations while preserving the direction and quality of aggregated model updates.

To address the above challenges, we propose {\bf SMELL}, a Sparsity-aware and Memory-Efficient federated fine-tuning framework for Long-context LLMs, which enables resource-constrained clients to collaboratively fine-tune long-context LLMs while balancing memory efficiency, model accuracy, and communication overhead. The key insight of SMELL is to integrate contextual token sparsity with forward-only optimization and federated coordination, jointly addressing the resource constraints and data heterogeneity inherent in federated long-context fine-tuning. Specifically, SMELL incorporates three complementary mechanisms: (i) Local Sparse-Forward Adaptation (LSFA), which combines contextual token sparsification with backpropagation-free gradient estimation to reduce local memory consumption while maintaining reliable model updates; (ii) Consensus-Anchored Sparsity Coordination (CASC), which establishes partial global consensus over token importance while preserving client-specific selection flexibility under non-IID data; and (iii) Communication-Efficient Progressive Update Recovery (CPUR), which exploits repeated gradient estimation to enable lightweight compressed transmissions and progressive reconstruction of model updates. By tightly integrating these mechanisms, SMELL provides a unified approach to memory-efficient local adaptation, heterogeneity-aware sparsity coordination, and communication-efficient federated model updating.

The main contributions are summarized as follows:
\begin{itemize}[leftmargin=*]
    \item We propose SMELL, a unified sparsity-aware federated fine-tuning framework on resource-constrained devices. To the best of our knowledge, SMELL is the first framework to jointly address memory-efficient local adaptation, cross-client sparsity coordination, and communication-efficient model updating in federated long-context fine-tuning.
    \item We develop LSFA and CASC mechanisms to integrate memory-efficient sparse forward computation with federated sparsity coordination. They reduce local memory overhead while balancing global token-importance consensus and client-specific selection flexibility under non-IID data.
    \item We design the CPUR mechanism with lightweight compressed transmissions and progressive update reconstruction, reducing individual communication payloads while preserving approximate model update directions.
    \item We conduct extensive experiments to evaluate SMELL. Results demonstrate up to 88.35\% and 13.25\% reductions in peak memory compared with backpropagation-based and forward-only fine-tuning, respectively, at a 16K context length, together with improvements in fine-tuning performance and effective compressed update recovery.
\end{itemize}

\section{Background and Motivation}

\subsection{Federated Fine-tuning of LLMs}

\begin{figure}[tbp]
    \centerline{\includegraphics[width=\linewidth]{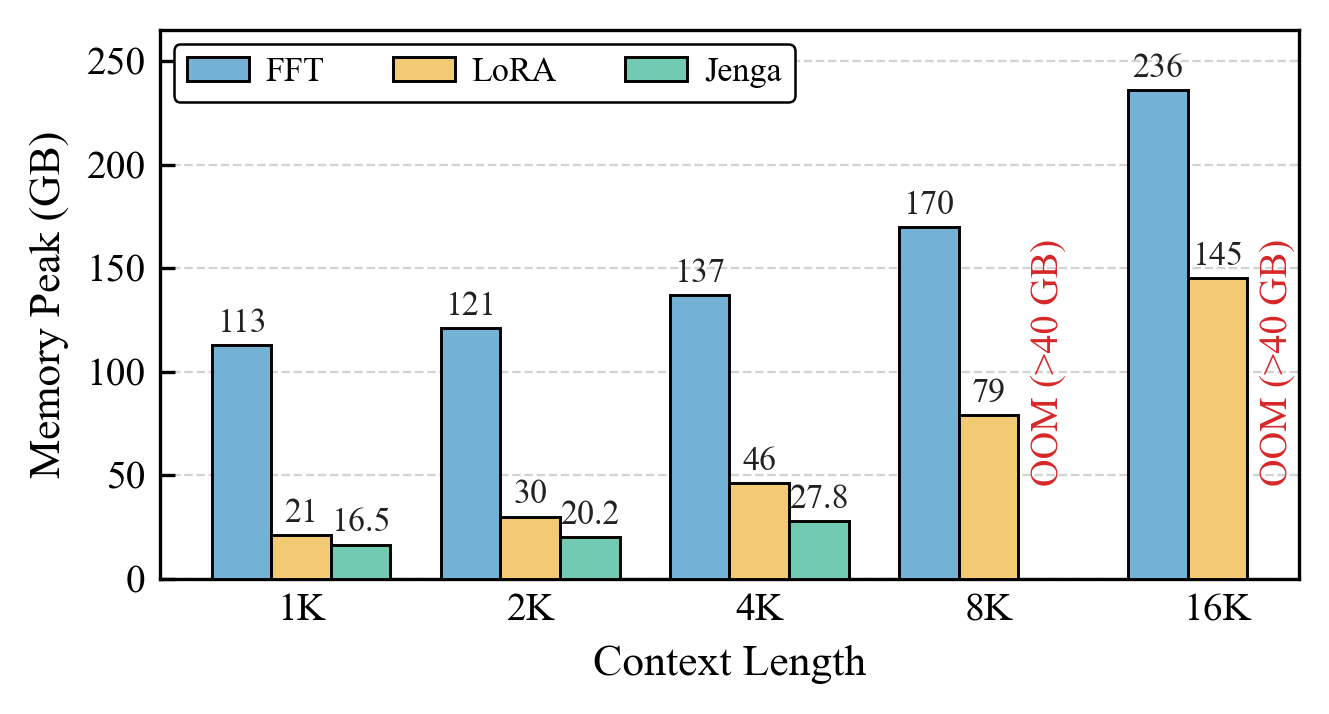}}
    \vspace{-0.2in}
    \caption{Peak memory of full fine-tuning, LoRA and Jenga on LLaMA2-7B.}
    \label{fig:memory_consumption}
    \vspace{-0.1in}
\end{figure}

Fine-tuning, which adapts pre-trained LLMs to diverse downstream applications, is essential for their effective deployment. In particular, fine-tuning on longer text sequences, together with appropriate context-extension techniques, enables LLMs to adapt beyond their pre-defined context windows and better support long-context applications, such as long document understanding and multi-turn conversations \cite{chen2023extendingcontextwindowlarge, chen2023longlora, yang2023longqlora, ding2024longrope}. However, the domain-specific data required for fine-tuning are often distributed across multiple devices and may contain sensitive information, making centralized data collection impractical due to privacy concerns. Federated fine-tuning of LLMs \cite{Cai2023Federated, he2020fedmlresearchlibrarybenchmark, cai2023fedadapterefficientfederatedlearning} provides a promising solution by enabling multiple devices to collaboratively adapt a shared model while keeping their raw training data local.

Due to the substantial memory requirements of LLM fine-tuning, deploying FedLLM on resource-constrained devices remains challenging. To improve fine-tuning efficiency, PEFT has emerged as a promising approach by updating only a small subset of parameters while preserving most of the pretrained model. A representative technique is LoRA, which freezes the pretrained model weights and introduces trainable low-rank matrices into selected Transformer layers. By substantially reducing the number of trainable parameters, LoRA lowers the memory overhead associated with gradients and optimizer states, which can collectively exceed the storage required for model weights during full-parameter fine-tuning. Consequently, LoRA and related PEFT techniques have been widely adopted in federated settings to improve the feasibility of distributed model adaptation. However, reducing trainable parameters does not directly resolve the activation-memory bottleneck, as processing long sequences still requires substantial intermediate activations for backpropagation, with memory consumption increasing rapidly with context length. For instance, fine-tuning LLaMA2-7B with a 16K context length requires an estimated peak memory of approximately 145 GB even with LoRA, far exceeding the 16–32 GB memory budgets of typical resource-constrained devices. This persistent gap between the memory demands of long-context fine-tuning and the limited resources available on federated devices highlights the need for training approaches that reduce activation memory beyond conventional parameter-efficient adaptation.

\subsection{Preliminary Experiments}

We conduct preliminary experiments along three dimensions: (i) the memory breakdown of long-context federated fine-tuning and the feasibility of sparsity-aware methods under device memory budgets; (ii) the reliability and resampling overhead of BP-free gradient estimation; and (iii) the global--local tension in sparse token selection under non-IID data.

\label{sec:mem_consumption}
{\bf Analysis 1: Federated fine-tuning memory breakdown.}
Prior studies, particularly Jenga~\cite{wang2025jenga}, have demonstrated that activation memory dominates long-context fine-tuning and that contextual token sparsification can substantially reduce this overhead, as shown in Fig.~\ref{fig:memory_consumption}. The peak memory comprises parameters, gradients, optimizer states, and activations; the activation term follows Sun et al.~\cite{HeimdaLLM2026} and is given by $A = 2L[(V_{fwd}+V_{bwd})L_{ctx}B d_{\text{model}}+5HL_{ctx}B]$, with notation in Table~\ref{tab:parameters}. However, these findings do not directly establish how sparsity-aware computation interacts with forward-only optimization and heterogeneous federated training.

\begin{figure}[tbp]
    \centerline{\includegraphics[width=\linewidth]{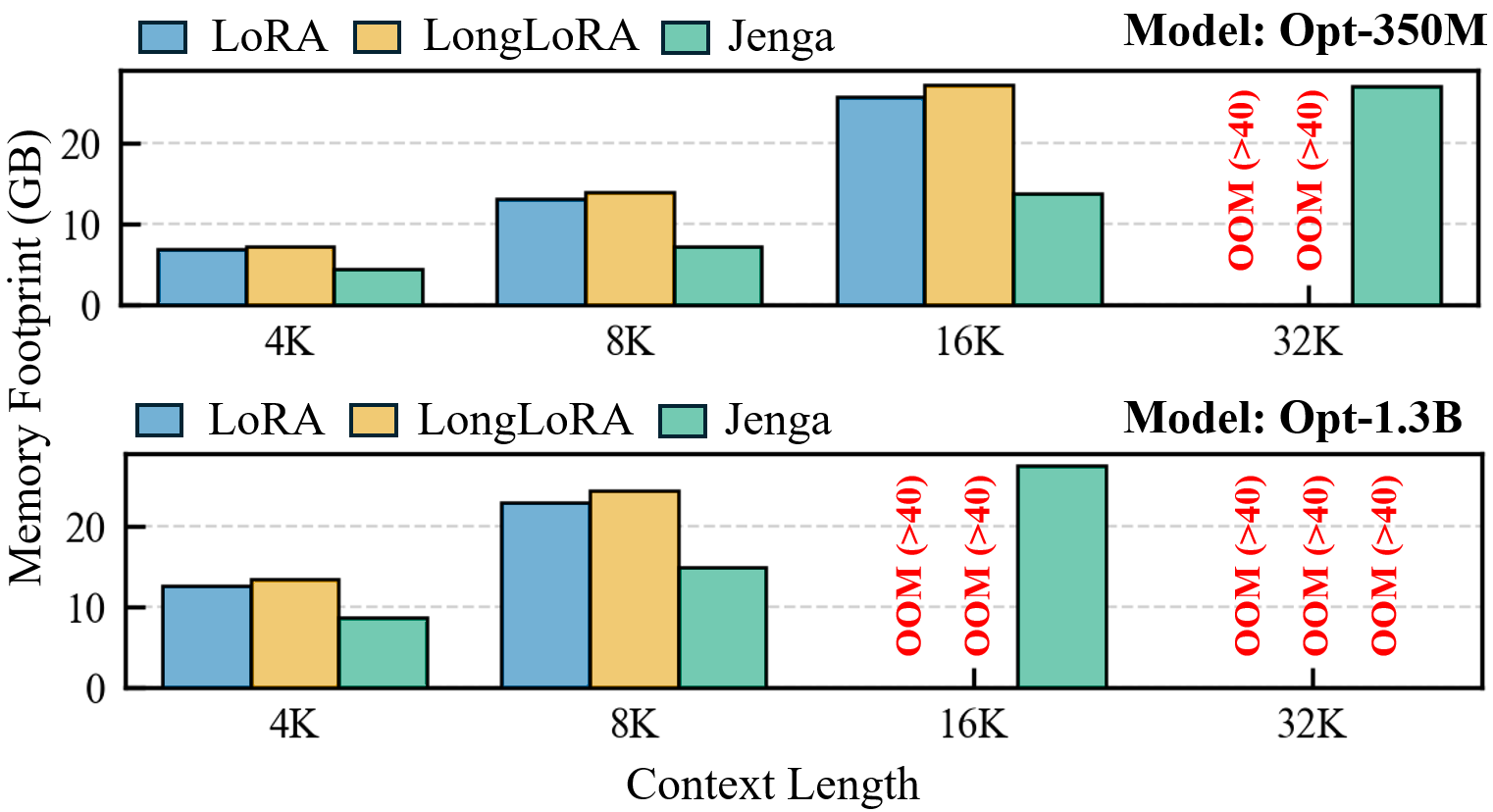}}
    \vspace{-0.15in}
    \caption{Peak memory of LoRA, LongLoRA and Jenga on OPT-350M (top) and OPT-1.3B (bottom).}
    \label{fig:jenga_infeasible}
    \vspace{-0.15in}
\end{figure}

In federated settings, device memory is even more constrained: we evaluate LoRA~\cite{hu2022iclr-lora}, LongLoRA~\cite{chen2023longlora}, and Jenga on OPT-350M and OPT-1.3B~\cite{zhang2022optopenpretrainedtransformer} at 16K and 32K, as depicted in Fig.~\ref{fig:jenga_infeasible}. At 16K, LoRA and LongLoRA exceed 40~GB on OPT-1.3B, while Jenga peaks at about 29~GB; at 32K, LoRA and LongLoRA also exceed 40~GB on OPT-350M, and all three methods exceed 40~GB on OPT-1.3B. \textit{Therefore, even with sparsification, existing approaches still cannot bring longer-context fine-tuning into the device memory budget; the next observation examines whether BP-free ZOO methods are feasible in long-context settings.}

\label{sec:motivation_zoo}

{\bf Analysis 2: Gradient estimation reliability and resampling overhead.}
Although BP-free ZOO-based methods substantially reduce activation memory, they rely on stochastic gradient estimates obtained through repeated perturbed forward passes. Improving estimation reliability may therefore require additional sampling, introducing a trade-off between update quality and computational overhead; this trade-off is particularly important for long-context fine-tuning, where each forward evaluation is computationally expensive.

\begin{table}[!t]
    \renewcommand{\arraystretch}{1.06}
    \caption{Description of major notations}
    \label{tab:parameters}
    \centering
    \small
    \begin{tabular}{l|p{6.5cm}}
    \hline
    \bfseries Symbol & \bfseries Definition\\
    \hline\hline
    $P,L$ & trainable parameters and transformer layers\\\hline
    $V_{fwd}$, $V_{bwd}$ & forward/backward activations per layer\\\hline
    $L_{ctx},d_{\text{model}}$ & context length and hidden dimension\\\hline
    $H,B,A$ & attention heads, batch size, peak activation memory\\\hline
    $\rho,\theta,\mathcal{L}$ & block keep ratio, LoRA parameters, training loss\\\hline
    $\varepsilon,v_j$ & finite-difference step, $j$-th perturbation direction\\\hline
    $D$, $D_{\{\cdot\}}$ & number of directions and its bounds\\\hline
    $g_{v_j}, \bar{g}_1$, $\bar{g}_2$ & gradient estimate and half-group mean estimates\\\hline
    $\mathrm{var}$, $T$ & half-mean variance and its threshold\\\hline
    $\eta$, $\gamma$, r& learning rate, clipping threshold, anchor ratio\\\hline
    $n_k$, $\Delta_k$ & samples and update of device $k$\\\hline
    $K,x$ & number of participating devices, update vector\\\hline
    $\bar{s}_{k,i},S_i$ & importance of block $i$ at $k$, aggregated importance\\\hline
    $\mathcal{A}_i$ & consensus anchor score of block $i$\\\hline
    $d_c$, $c$, $\kappa$ & buckets per channel, channels, rate ($d_c=\kappa d$)\\\hline
    $h_j$, $\sigma_j$ & hash and sign mappings of channel $j$\\\hline
    $a_j$, $b_j$, $M$ & hash/sign constants of channel $j$\\\hline
    $z_j$, $\hat{x}$ & compressed channel and recovered update\\
    \hline
    \end{tabular}
\end{table}

\begin{figure}[htbp]
    \centerline{\includegraphics[width=0.5\linewidth]{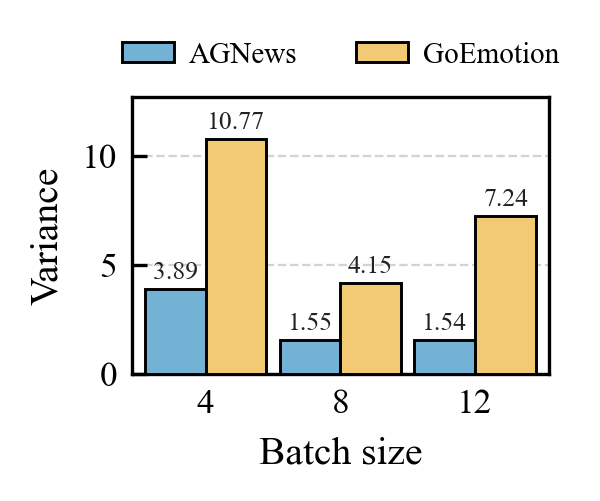}\includegraphics[width=0.5\linewidth]{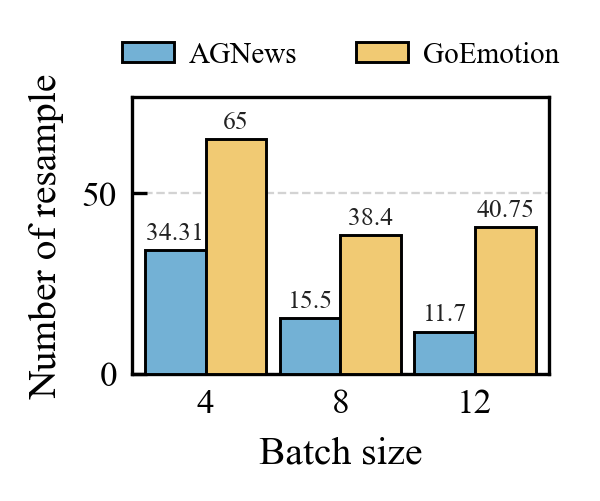}}
    \vspace{-0.15in}
    \caption{ZOO gradient variance (left) and the required number of resampling steps (right) across datasets and batch sizes.}
    \label{fig:var_res}
\end{figure}

To investigate this issue, we examine the gradient-estimation variance and resampling overhead of FwdLLM~\cite{FwdLLM} using DistilBERT~\cite{sanh2020distilbertdistilledversionbert} with adapters~\cite{Pfeiffer2020AdapterHubAF}. We compare AGNews~\cite{Zhang2015CharacterlevelCN}, with a maximum sequence length of 64, and GoEmotion~\cite{demszky2020goemotionsdatasetfinegrainedemotions}, constructed from concatenated few-shot examples with a maximum sequence length of 256. As shown in Fig.~\ref{fig:var_res}, the longer-sequence workload exhibits higher empirical gradient variance and requires more resampling operations across the evaluated batch sizes. For matched batch sizes, the peak number of resampling operations on GoEmotion is about 1.9--3.5$\times$ that of AGNews. Although the two workloads also differ in their data characteristics, these results highlight the potential computational overhead of obtaining reliable zeroth-order updates. \textit{This observation suggests that eliminating backpropagation alone is insufficient for efficient long-context fine-tuning: the sampling budget must also be carefully considered to balance gradient estimation reliability and training efficiency.}

\begin{figure*}[tbp]
    \centerline{\includegraphics[width=1\linewidth]{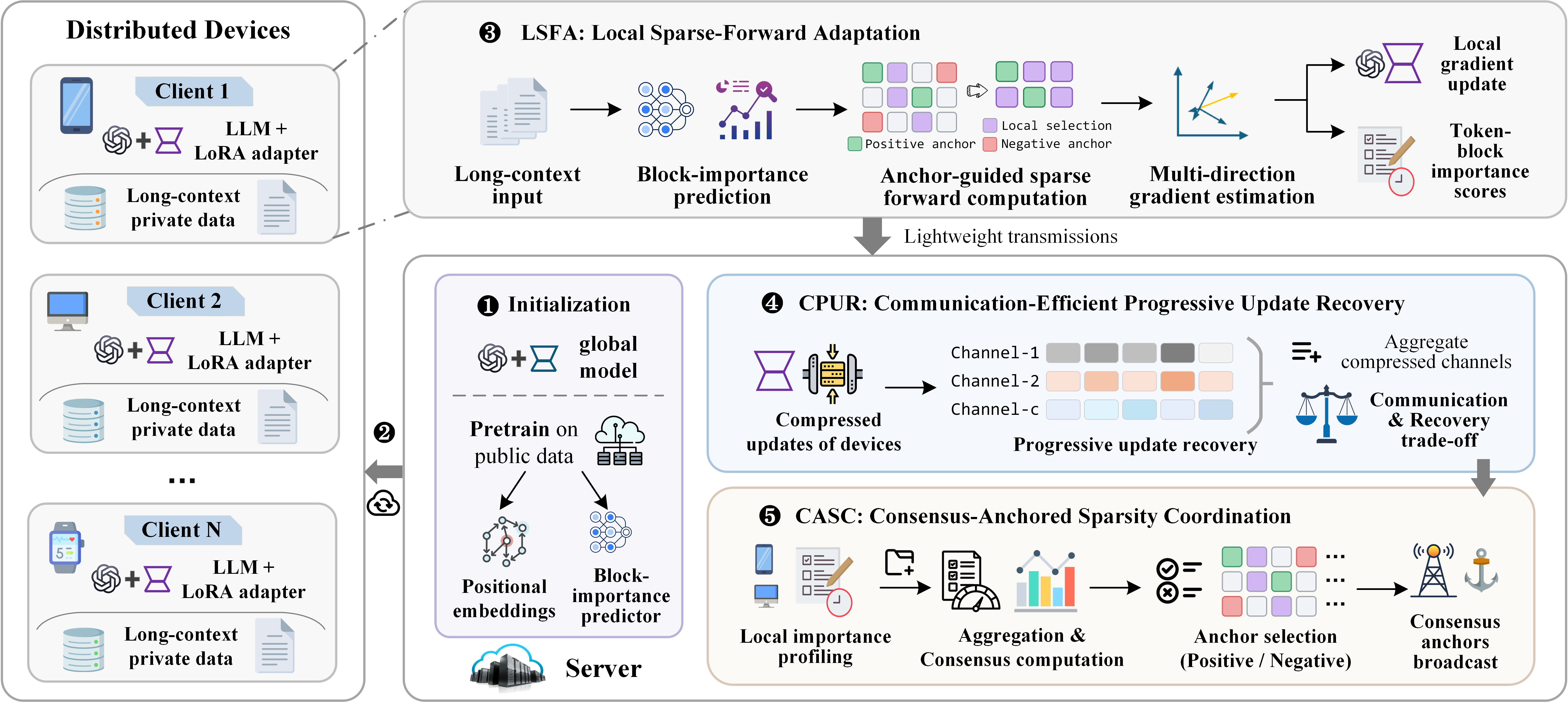}}
    \vspace{-0.1in}
    \caption{Overview of the SMELL framework, comprising three key mechanisms: LSFA, CASC, and CPUR.}
    \label{fig:overview}
    \vspace{-0.1in}
\end{figure*}
\label{sec:motivation_non_iid}

{\bf Analysis 3: The global--local sparsity dilemma in Federated Fine-tuning.}
Contextual token sparsification identifies informative token blocks based on input-dependent importance scores and has demonstrated effectiveness in centralized settings~\cite{wang2025jenga}. However, extending this approach to federated fine-tuning introduces a new challenge: heterogeneous data distributions across devices may lead to divergent token-importance patterns and, consequently, distinct sparse computation paths. Although such heterogeneity does not inherently impede model aggregation, it causes local updates to be derived from different subsets of contextual information. Purely local token selection may fail to capture importance patterns shared across devices, whereas imposing a globally uniform sparse mask risks discarding information essential to devices. This gives rise to a fundamental global--local coordination problem: how to leverage shared token-importance patterns while preserving device-specific sparsity preferences.

To investigate this trade-off, we conduct experiments on LLaMA2-7B~\cite{Touvron2023Llama2} using WikiText~\cite{merity2016pointer}, with data partitioned across devices according to a Dirichlet distribution. We vary the number of devices \(K\!\in\!\{10,20,50\}\) and the Dirichlet concentration parameter \(\alpha\!\in\!\{0.1,0.2,0.5\}\), where smaller $\alpha$ indicates greater data heterogeneity. We then compare locally selected token masks against a centralized reference mask using the mask Intersection Rate (IR) and the Perplexity Difference ($\Delta$PPL), defined as follows:
\begin{IEEEeqnarray}{rCl}
    \mathrm{IR} &=& {|\mathcal{M}_{\text{local}}\cap\mathcal{M}_{\text{central}}|}/|\mathcal{M}_{\text{central}}|,\\
    \Delta\mathrm{PPL} &=& \mathrm{PPL}(\mathcal{M}_{\text{local}})-\mathrm{PPL}(\mathcal{M}_{\text{central}}).\label{eq:ir_drift}
\end{IEEEeqnarray}

Through preliminary experiments, IR between locally selected masks and the centralized reference mask ranges from 0.20 to 0.46, revealing substantial discrepancies in token selection under non-IID data distributions. Meanwhile, the corresponding $\Delta$PPL varies from $-0.0369$ to $0.15$ (average $0.077$). In 9 experiments, 8 $\Delta$PPL of them are positive, which means non-IID data may be less effective for federated learning. Notably, with 10 devices and $\alpha=0.5$, the IR drops to 0.20, yet $\Delta$PPL remains negative ($-0.0369$), suggesting that locally selected masks can outperform the centralized reference despite limited overlap. These observations reveal that mask agreement alone is insufficient to characterize the effectiveness of sparse computation. Rather than enforcing a globally uniform mask or relying exclusively on independent local selections, \textit{federated fine-tuning calls for a coordination strategy that captures shared token-importance patterns while preserving device-specific selection flexibility.}

These observations reveal three interrelated challenges in federated long-context LLM fine-tuning: substantial memory demands driven by long-context activations, the trade-off between gradient estimation reliability and resampling overhead, and the need to balance global sparsity consensus with device-specific selection flexibility under non-IID data. Addressing these challenges requires more than isolated optimizations. Instead, it calls for a unified framework that integrates memory-efficient sparse computation, reliable forward-only adaptation, and cross-device sparsity coordination, while enabling efficient communication of model updates.

\section{Framework Design}

\subsection{Overview of SMELL}

We propose SMELL, a sparsity-aware and memory-efficient federated fine-tuning framework for Long-context LLMs on resource-constrained devices. \textit{Its key insight is to exploit contextual token sparsity as a foundation for resource-efficient federated training, while jointly addressing the optimization reliability and cross-device coordination challenges introduced by sparse computation under non-IID data.} Rather than treating memory reduction, heterogeneous token selection, and communication efficiency as isolated problems, SMELL integrates them into a unified federated training pipeline to balance resource efficiency and model performance. Fig.~\ref{fig:overview} presents an overview of our framework, which is built upon three mutually coupled mechanisms:

{\bf Local Sparse-Forward Adaptation (Sec.~\ref{sec:LSFA}).} LSFA integrates contextual token sparsification with backpropagation-free parameter adaptation to jointly reduce attention-related and backward activation memory. It further leverages multi-direction forward-gradient estimation to improve update reliability under constrained device resources.

{\bf Consensus-Anchored Sparsity Coordination (Sec.~\ref{sec:CASC}).} CASC coordinates heterogeneous token-importance patterns across non-IID devices by partial global consensus. By anchoring globally preferred token blocks while preserving local selection flexibility, it balances shared sparsity preferences with device-specific information needs.

{\bf Communication-Efficient Progressive Update Recovery (Sec.~\ref{sec:CPUR}).} CPUR exploits repeated stochastic gradient estimation to progressively reconstruct model updates from compressed representations. By transmitting lightweight channels across sampling rounds, it reduces individual uplink payloads while approximating the aggregated update direction.

Together, these mechanisms enable SMELL to integrate memory-efficient local computation, heterogeneity-aware sparsity coordination, and compressed model updates within a unified federated fine-tuning framework. The server performs model aggregation and sparse-anchor coordination, while the devices retain their private training data and execute the computationally intensive fine-tuning operations locally. The detailed workflow is described as follows:

{\bf Framework Workflow:} SMELL consists of a central server and multiple distributed devices, each equipped with an identically initialized LLM and LoRA adapters. \ding{182} Initialization: Before federated training, the server initializes the global model and pretrains the positional embeddings and block-importance predictor on public data, followed by broadcasting the initialized model to all devices. \ding{183} Global synchronization: At the beginning of each round, the server distributes the global LoRA parameters, shared random seeds, and consensus sparsity anchors to participating devices. \ding{184} Local adaptation: Each device performs LSFA on its private long-context data, integrating anchor-guided token sparsification with backpropagation-free gradient estimation to obtain memory-efficient local updates. \ding{185} Compressed update transmission: CPUR progressively transmits compact gradient representations and accumulates them at the server to recover approximate update directions. The server also monitors gradient-estimation consistency to determine whether additional sampling is required. \ding{186} Sparsity coordination and global aggregation: Devices upload locally averaged token-block importance scores, enabling CASC to identify shared sparsity preferences and construct consensus anchors while preserving device-specific selection flexibility. The server then aggregates the clipped model updates and broadcasts the updated LoRA parameters and sparsity anchors for the next round. Through this iterative workflow, SMELL integrates memory-efficient local adaptation, cross-device sparsity coordination, and compressed update aggregation without sharing raw device data.

\subsection{Local Sparse-Forward Adaptation Mechanism}
\label{sec:LSFA}

{\bf Design rationale.} To enable long-context fine-tuning on memory-constrained devices, we develop the {\bf LSFA} mechanism, which integrates contextual token sparsification with backpropagation-free optimization into a unified local training procedure. While existing forward-only methods eliminate the need to store activations for backpropagation, they still incur substantial memory overhead from long-context attention. Conversely, token sparsification reduces attention-related memory consumption but does not eliminate the need for backward propagation. By combining these complementary approaches, LSFA performs gradient estimation directly through sparsified forward passes, jointly reducing attention-related and backpropagation-related memory overhead while enabling efficient LoRA adaptation.

{\bf Anchor-guided sparse forward computation.} LSFA first constructs a sparse computation path for each local input. Following the block-level sparsification design of Jenga \cite{wang2025jenga}, the input sequence is divided into non-overlapping token blocks of size 64. For each sparsified Transformer layer, a pretrained MLP-based importance predictor assigns scores to token blocks using the hidden states before the attention operation. The resulting scores are combined with the consensus anchor scores received from the server through CASC (Sec.~\ref{sec:CASC}). Blocks assigned positive or negative infinite anchor scores are forced to be retained or discarded, respectively, while the remaining blocks are ranked according to their local importance scores. Given a keep ratio $\rho$ (the fraction of token blocks retained in each sparsified layer), only the top-ranked blocks are retained for the corresponding attention computation, avoiding the materialization of KV states for discarded blocks. Note that, sparsification is applied to the upper Transformer layers, while the lower layers and the final layer remain dense. For example, with a 16K-token input and $\rho = 0.4$, LSFA retains 102 out of 256 token blocks in each sparsified layer. This design preserves input-dependent local selection while incorporating the shared sparsity preferences established through federated coordination.

{\bf Forward-only parameter adaptation. }On top of sparse forward computation, LSFA estimates local LoRA gradients without backpropagation. Let $\theta$ denote the trainable LoRA parameters, while the pretrained backbone remains frozen. For a local sparse training loss $\mathcal{L}(\theta)$, the directional derivative along a perturbation vector $v$ with step size $\varepsilon$ is defined as:

{\setlength{\abovedisplayskip}{-6pt}
	\setlength{\belowdisplayskip}{3pt}
	\begin{eqnarray}
    \nabla_{v}\mathcal{L}(\theta) = \lim_{\varepsilon \to 0} \frac{\mathcal{L}(\theta + \varepsilon v) - \mathcal{L}(\theta - \varepsilon v)}{2\varepsilon} = \nabla \mathcal{L}(\theta) \cdot v.
    \vspace{-0.05in}
   \end{eqnarray}

Rather than explicitly computing the gradient through backward propagation, LSFA samples independent perturbation directions ($v_j\sim\mathcal N(0,I)$) and estimates the gradient through central finite differences:

{\setlength{\abovedisplayskip}{-6pt}
	\setlength{\belowdisplayskip}{3pt}
	\begin{eqnarray}
    \label{eq:forward_gradient}
    g_{v_j}(\theta) := \frac{\mathcal{L}(\theta + \varepsilon v_j) - \mathcal{L}(\theta - \varepsilon v_j)}{2\varepsilon}\, v_j,
\end{eqnarray}
where $g_{v_j}(\theta)$ denotes the forward-gradient estimate along the $j$-th perturbation direction $v_j$, and both loss evaluations follow the anchor-guided sparse forward computation described above. By averaging $D$ independent estimates, where $D$ denotes the number of perturbation directions sampled per round, LSFA obtains the local update direction:

{\setlength{\abovedisplayskip}{-6pt}
	\setlength{\belowdisplayskip}{3pt}
	\begin{eqnarray}
    \nabla \hat{\mathcal{L}}(\theta) = \frac{1}{D}\sum\nolimits_{j=1}^D g_{v_j}(\theta).
\end{eqnarray}

Under independent perturbation sampling, averaging reduces the variance of the stochastic gradient estimator as $D$ increases. This improves estimation stability at the cost of additional forward evaluations. Since each evaluation operates on sparsified attention computations, LSFA combines memory-efficient forward execution with repeated gradient estimation without requiring backpropagation.

{\bf Gradient estimation consistency.} To characterize the reliability of the estimated update direction without accessing the true gradient, we further consider the consistency between independent groups of gradient estimates. Specifically, the $D$ estimates are divided chronologically into two equal-sized groups; let $\bar{g}_1,\bar{g}_2\in\mathbb{R}^d$ denote the mean gradient estimates of the first and second groups, respectively, where $d$ is the dimension of the trainable gradient vector. Their discrepancy, after normalization, is measured by

{\setlength{\abovedisplayskip}{-6pt}
	\setlength{\belowdisplayskip}{3pt}
    \begin{eqnarray}
    \label{eq:var}
    \mathrm{var} &:=& \frac{1}{2d}\left\|\bar g_1 - \bar g_2\right\|_2^2.
\end{eqnarray}
A smaller discrepancy indicates greater agreement between the two groups of stochastic estimates, providing an observable measure of estimation consistency.

After local computation, the estimated gradient updates are passed to CPUR for compressed transmission, while the token-block importance scores are averaged over local samples and provided to CASC for cross-device sparsity coordination. The updated consensus anchors subsequently guide the next round of local sparse forward computation. Through this integration, LSFA forms the memory-efficient computational foundation of SMELL, connecting local forward-only adaptation with federated sparsity coordination and compressed update aggregation.

\subsection{Consensus-Anchored Sparsity Coordination Mechanism}
\label{sec:CASC}

{\bf Design rationale.} While LSFA enables memory-efficient sparse computation on individual devices, independently selecting token blocks under non-IID data distributions may lead to heterogeneous sparsity patterns across devices. Since local model updates are generated from different subsets of contextual information, purely local sparsification may overlook token-importance patterns shared across the federation. Conversely, enforcing an identical sparse mask across all devices may discard information particularly important to individual data distributions. To balance these competing requirements, we introduce the {\bf CASC} mechanism, which establishes partial global consensus over token importance while preserving device-specific selection flexibility.

{\bf Local importance profiling.} During LSFA, each device evaluates token-block importance using a pretrained predictor. Rather than transmitting raw training data or token representations, CASC reuses the importance scores generated during sparse forward computation. Specifically, for each sparsified layer, device $k$ averages the importance scores at each token-block position $i$ across its local training samples, yielding a compact importance profile $\bar{s}_{k,i}$. These profiles are then uploaded to the server for cross-device sparsity coordination.

{\bf Global consensus anchoring.} Upon receiving the local importance profiles, the server aggregates the scores of corresponding token-block positions across participating devices: $S_i=\sum_{k=1}^{K}\bar{s}_{k,i}$, where $K$ denotes the number of participating devices and $S_i$ represents the aggregated importance score of block position $i$. This aggregation is performed independently for each sparsified transformer layer. Based on the resulting global ranking, CASC constructs a consensus anchor score $\mathcal{A}_i$ for each block using the following rule:

{\setlength{\abovedisplayskip}{-6pt}
	\setlength{\belowdisplayskip}{3pt}
    \begin{eqnarray}
    \label{eq:CASC}
\mathcal{A}_i=\begin{cases}
+\infty,&\text{$i$ ranks in the top $r$ fraction;}\\
-\infty,&\text{$i$ ranks in the bottom $r$ fraction;}\\
0,&\text{otherwise},
\end{cases}
\end{eqnarray}
where $r$ is the anchor ratio controlling the extent of cross-device coordination. Blocks with the highest aggregated importance are designated as positive anchors, while those with the lowest importance become negative anchors. The remaining blocks are left unconstrained, allowing their selection to adapt to local data characteristics.

{\bf Consensus-guided local sparsification.} At the beginning of the next federated round, the server broadcasts the consensus anchor scores to participating devices. During LSFA, each device incorporates these anchors into its locally predicted importance scores before selecting token blocks. Positive anchors are always retained, negative anchors are excluded, and unconstrained blocks are selected according to their local importance rankings under the prescribed sparsity budget $\rho$. For instance, with 256 token blocks, $r=0.2$, and $\rho=0.4$, approximately 51 blocks are globally retained and 51 are globally excluded, while the device selects the remaining 51 retained blocks according to its local importance scores. This partial anchoring strategy enables our CASC mechanism to incorporate shared importance preferences without imposing a fully identical sparse mask on every device.

By iteratively aggregating local importance profiles and broadcasting consensus anchors, CASC establishes a feedback loop between global sparsity coordination and local sparse-forward adaptation. Rather than enforcing uniform masks or relying solely on independent local selection, CASC balances shared sparsity preferences with device-specific flexibility through partial consensus anchoring. Moreover, this coordination requires only block-level importance statistics and lightweight server-side ranking, without sharing raw device data or performing additional server-side model training.

\subsection{Communication-Efficient Progressive Update Recovery Mechanism}
\label{sec:CPUR}

{\bf Design rationale.} While LSFA reduces the memory overhead of local long-context fine-tuning, federated training still requires devices to communicate their estimated LoRA updates to the server. Although parameter-efficient adaptation reduces the number of trainable parameters, transmitting high-dimensional updates can remain costly, particularly for larger models and bandwidth-constrained devices. Moreover, forward-only optimization relies on repeated stochastic gradient estimation, creating opportunities to coordinate update representation with the sampling process. To exploit this opportunity, we introduce {\bf CPUR} mechanism, which integrates lightweight update encoding with progressive server-side reconstruction. Instead of transmitting a complete high-dimensional update in each communication round, CPUR distributes compressed representations across successive sampling rounds, allowing the server to accumulate complementary information for approximate update recovery.

{\bf Compact update encoding.} CPUR builds on the Count-Sketch principle \cite{rothchild2020fetchsgdcommunicationefficientfederatedlearning} to encode high-dimensional model updates into fixed-size representations. Consider an estimated LoRA update vector $x\in\mathbb{R}^d$. Directly transmitting this vector in FP32 requires $4d$ bytes. CPUR instead constructs independent compressed channels, each containing $d_c$ buckets, where $d_c<d$. For each channel $j$, two seed-defined mappings are employed: a hash function $h_j(p)$, which maps each parameter index $p\in\{0,\dots,d-1\}$ of the update vector $x$ to a bucket $\{0,\dots,d_c-1\}$, and a sign function $\sigma_j(p)$, which determines the sign of each contribution:

{\setlength{\abovedisplayskip}{-6pt}
	\setlength{\belowdisplayskip}{0pt}
    \begin{eqnarray}
    h_j(p) := \mathrm{round}(p \cdot a_j) \bmod d_c,\ a_j \in [2, 10^6).
\end{eqnarray}
{\setlength{\abovedisplayskip}{0pt}
	\setlength{\belowdisplayskip}{3pt}
    \begin{eqnarray}
    \sigma_j(p) := \begin{cases}
        +1, & \text{$(p \cdot b_j \bmod M) \le M/2$;}\\
        -1, & \text{otherwise.}
    \end{cases}
\end{eqnarray}

The compressed representation is obtained by accumulating signed update elements in their corresponding hash buckets. Specifically, for channel $j$, the compressed vector $z_j\in\mathbb{R}^{d_c}$ has its $b$-th entry defined as follows:

{\setlength{\abovedisplayskip}{-6pt}
	\setlength{\belowdisplayskip}{3pt}
    \begin{eqnarray}
    z_j[b] = \sum\nolimits_{p:\ h_j(p)=b} x_p \cdot \sigma_j(p),\ b\in\{0,\dots,d_c-1\}.
\end{eqnarray}

Each channel therefore requires transmitting only $d_c$ numerical values instead of the original $d$, reducing the payload of an individual transmission by a factor of $d/d_c$. The random seeds are shared between devices and the server, allowing the compressed representation to be decoded without transmitting the complete mapping tables.

{\bf Progressive update recovery.} The distinctive feature of CPUR lies in how these compressed channels are incorporated into the repeated update-estimation process of LSFA. During successive resampling communication rounds, devices transmit compact channels generated using different random mappings. Rather than requiring each transmission to independently reconstruct the complete update, the server progressively accumulates the received channels and combines their information to estimate the update direction.

By aggregating \(c\) compressed channels, the recovered update is $\hat{x}$, whose $p$-th entry is denoted by $\hat{x}_p$ and computed as

{\setlength{\abovedisplayskip}{-6pt}
	\setlength{\belowdisplayskip}{3pt}
    \begin{eqnarray}
    \label{eq:decompress}
    \hat{x}_p = \frac{1}{c}\sum\nolimits_{j=0}^{c-1} z_j\!\left[h_j(p)\right] \cdot \sigma_j(p).
\end{eqnarray}

Each channel provides an independent, potentially noisy estimate, while their aggregation is intended to reduce the interference introduced by hash collisions. Consequently, CPUR decouples the size of an individual uplink transmission from the amount of information accumulated for update recovery. The reconstructed updates can then be incorporated into the server-side federated aggregation process.

{\bf Recovery properties.} To characterize the reconstruction process, we let $\hat{x}^{j}$ denote the update vector encoded in channel $j$. Then, we expand Eq.~(\ref{eq:decompress}) and get

{\setlength{\abovedisplayskip}{-6pt}
	\setlength{\belowdisplayskip}{3pt}
    \begin{eqnarray}
    \hat{x}_p = \frac{1}{c}\sum\nolimits_{j=0}^{c-1}\sum\nolimits_{q=0}^{d-1} x_q\, \sigma_j(p)\, \sigma_j(q)\, \mathbf{1}\{h_j(q)=h_j(p)\},
\end{eqnarray}
where $q\in\{0,\dots,d-1\}$ is the parameter index of the update vector (the same index set as $p$), and $\mathbf{1}\{\cdot\}$ denotes the indicator function, which equals $1$ when the condition inside the braces holds and $0$ otherwise. The indicator therefore retains only the positions $q$ that collide with $p$ in channel $j$.

We assume that, for each channel $j$, $h_j$ is drawn uniformly from a
pairwise independent hash family, the signs $\sigma_j(p)$ are four-wise
independent Rademacher variables~\cite{CHARIKAR20043}, and $h_j$ is independent of $\sigma_j$;
different channels use independent seeds. These assumptions imply
$\mathbb{E}[\sigma_j(p)\sigma_j(q)]=0$ for all $p\ne q$, so the cross-element
interference terms vanish in expectation and $\mathbb{E}[\hat{x}_p]=x_p$.
Consequently, conditioned on the input update vectors, we have the following
equation:

{\setlength{\abovedisplayskip}{-6pt}
	\setlength{\belowdisplayskip}{3pt}
    \begin{eqnarray}
    \mathbb{E}[\hat{x}_p]\!=\!\!\frac{1}{c}\sum_j \!x_p\!+\!\frac{1}{c}\sum_j\!\!\sum_{q\ne p} \!x_q\, \mathbb{E}[\sigma_j(p)\sigma_j(q)]\, \mathbf{1}\{h_j(q)\!\!=\!\!h_j(p)\}. \nonumber
\end{eqnarray}

This indicates that progressive recovery can target the average of the encoded stochastic updates, even when different channels represent different gradient estimates, provided the required randomization assumptions hold. Under the same assumptions, Charikar et al.~\cite{CHARIKAR20043} give a corresponding conditional reconstruction-error bound as $\mathbb{E}\!\left[(\hat{x}_p-x_p)^2\right] \le \|x\|_2^2/(cd_c).$ Accordingly, for the fixed-vector case, Chebyshev's inequality gives, for any deviation coefficient $\zeta>0$,
$P\!\left(|\hat{x}_p - x_p| \ge \zeta \,\|x\|_2\right) \le 1/(\zeta^2 \, c \, d_c),$
i.e., $\zeta$ bounds the deviation $|\hat{x}_p-x_p|$ relative to the update norm $\|x\|_2$. These bounds characterize reconstruction error under the stated randomization assumptions. The above analysis assumes idealized randomized mappings. In our implementation, $h_j$ and $\sigma_j$ are instantiated by the simple linear hash and threshold sign functions defined above; these provide a lightweight approximation to the idealized assumptions. When CPUR operates on stochastic estimates generated across rounds, the overall optimization error additionally depends on the variability and potential bias of the forward-gradient estimates.

{\bf Communication--recovery trade-off.} CPUR provides flexible control over communication and reconstruction fidelity through the bucket count $d_c$ and the number of accumulated channels $c$. Reducing $d_c$ lowers the uplink payload per transmission, while accumulating more independent channels improves reconstruction fidelity. In our implementation, setting $d_c=d/8$ reduces each transmission to one-eighth of the full FP32 update size, with bf16 encoding further halving the payload. However, the cumulative communication volume scales with $c,d_c$, highlighting the trade-off between transmission efficiency and reconstruction accuracy.

By coupling compact update encoding with the repeated sampling structure of LSFA, CPUR provides a progressive communication and reconstruction procedure. Together with CASC, which coordinates sparse computation across heterogeneous devices, CPUR completes the framework's integration of memory-efficient local adaptation, federated sparsity coordination, and compressed model-update exchange.

\subsection{Pseudocode of SMELL}

Algorithm~\ref{alg:smell} summarizes the server-side protocol. The server initializes the global LoRA parameters $\theta$, the consensus anchors $\mathcal{A}$, and the aggregate importance vector $S$, and broadcasts the initialized model and the shared constants $(\rho,\varepsilon,\kappa)$ to all devices (Line 1-2). In each round, for every participating device $k$, it resets the direction counter $D_k$, the recovered-update set $G_k$, and the variance estimate $\mathrm{var}$, while the device-side history $H_k$ and profile $\bar{s}_k$ are empty at the start of the round (Lines 3-5). It then draws a seed $s$ and sends $(\theta,\mathcal{A},s)$ to device $k$ (Line 6). While the stopping condition is not met, the server calls $\textsc{DeviceUpdate}(\theta,\mathcal{A},s,\rho,\varepsilon,\kappa)$ at device $k$ and receives a compressed channel $z$ and the variance $\mathrm{var}$ (Line 7-8). The server decompresses $z$ into a recovered update $\hat{x}$, inserts $\hat{x}$ into $G_k$, and increments $D_k$ (Lines 9-10). The loop terminates when $D_k\ge D_{\min}$ and either $\mathrm{var}\le T$ or $D_k\ge D_{\max}$ (Line 7). The server then averages the recovered updates into $\Delta_k$ and clips it with threshold $\gamma$, while the device uploads its averaged profile $\bar{s}_k/D_k$ to the server (Line 12-14). After all devices finish, the server applies the sample-weighted FedAvg update, aggregates the uploaded profiles as $S=\sum_{k=1}^{K}\bar{s}_k$, and sets the top-$r$ and bottom-$r$ blocks of $S$ to $+\infty$ and $-\infty$ in each sparsified layer separately to form the next-round anchors according to Eq.~\ref{eq:CASC}; the remaining blocks are set to $0$ (Line 16-19). The server returns the final global LoRA parameters $\theta$ after $R$ training rounds (Line 20).
\setlength{\textfloatsep}{10pt plus 2pt minus 2pt}
\begin{algorithm}[t]
\caption{SMELL: Federated Long-Context Fine-tuning}
\label{alg:smell}
\small
\begin{algorithmic}[1]
\REQUIRE $R$ rounds, $K$ devices, $D_{\min},D_{\max}$ directions, threshold $T$, step size $\varepsilon$, learning rate $\eta$, keep ratio $\rho$, anchor ratio $r$, compression rate $\kappa$, clip threshold $\gamma$
\ENSURE global LoRA $\theta$
\STATE $\theta\leftarrow\theta_0$, $\mathcal{A}\leftarrow\mathbf{0}$, $S\leftarrow\mathbf{0}$;
\STATE server broadcasts the initialized model $\theta$ and the shared constants $(\rho,\varepsilon,\kappa)$ to all devices;
\FOR{$t=0$ to $R-1$}
    \FOR{each device $k=1,\dots,K$ in parallel}
        \STATE $D_k\leftarrow0$, $G_k\leftarrow\emptyset$, $\mathrm{var}\leftarrow\infty$;
        \STATE server draws seed $s$, sends $(\theta,\mathcal{A},s)$ to device $k$;
        \WHILE{$D_k<D_{\min}$ \textbf{or} ($\mathrm{var}>T$ \textbf{and} $D_k<D_{\max}$)}
            \STATE $(z,\mathrm{var})\leftarrow\textsc{DeviceUpdate}(\theta,\mathcal{A},s,\rho,\varepsilon,\kappa)$;
            \STATE $\hat{x}\leftarrow\textsc{CPUR}^{-1}(z)$;
            \STATE $G_k\leftarrow G_k\cup\{\hat{x}\}$, $D_k\leftarrow D_k+1$;
        \ENDWHILE
        \STATE $\Delta_k\leftarrow\frac{1}{D_k}\sum_{\hat{x}\in G_k}\hat{x}$;
        \STATE $\Delta_k\leftarrow\textsc{Clip}(\Delta_k,\gamma)$;
        \STATE device $k$ sends $\bar{s}_k/D_k$ to the server;
    \ENDFOR
    \STATE $\theta\leftarrow\theta-\eta\sum_{k=1}^{K}\frac{n_k}{\sum_{k'} n_{k'}}\Delta_k$;
    \STATE $S\leftarrow\sum_{k=1}^{K}\bar{s}_k$;
    \STATE $\mathcal{A}\leftarrow\textsc{CASC}(S)$;
\ENDFOR
\STATE \textbf{return} $\theta$;
\end{algorithmic}
\end{algorithm}

Algorithm~\ref{alg:device} details one \textsc{DeviceUpdate} call. Given $\theta$, $\mathcal{A}$, $s$, and the shared constants $(\rho,\varepsilon,\kappa)$, the device draws one perturbation direction $v\sim N(0,I)$ from $s$, samples a local input $\xi$, and computes its block scores $\omega$ (Line 1-3). It adds $\omega$ to its local profile $\bar{s}_k$ and builds the local sparse mask $\mathcal{M}=\textsc{BuildMask}(\omega,\mathcal{A},\rho)$ (Line 4-5). It then evaluates the loss at $\theta+\varepsilon v$ and $\theta-\varepsilon v$ under $\mathcal{M}$, and forms the forward-gradient estimate $g_v$ (Line 6). The estimate is appended to the device history $H_k$, which is empty at the start of each round, and the half-mean variance of $H_k$ gives $\mathrm{var}$ according to Eq.~\ref{eq:var} (Line 7). In the end, $g_v$ is compressed by CPUR at rate $\kappa$, and the resulting channel $z$ together with $\mathrm{var}$ is returned (Lines 8–9).
\setlength{\textfloatsep}{10pt plus 2pt minus 2pt}
\vspace{-6pt}
\begin{algorithm}[t]
\caption{DeviceUpdate: One Forward-Gradient Direction}
\label{alg:device}
\small
\begin{algorithmic}[1]
\REQUIRE global $\theta$, consensus anchors $\mathcal{A}$, seed $s$, shared constants $\rho$ (keep ratio), $\varepsilon$ (step size), $\kappa$ (compression rate), device history $H_k$ (empty at round start), device profile $\bar{s}_k$ (empty at round start)
\ENSURE compressed channel $z$, variance $\mathrm{var}$
\STATE draw one perturbation direction $v\sim N(0,I)$ from $s$;
\STATE sample a local input $\xi$;
\STATE $\omega\leftarrow$ block scores of $\xi$;
\STATE $\bar{s}_k\leftarrow\bar{s}_k+\omega$;
\STATE $\mathcal{M}\leftarrow\textsc{BuildMask}(\omega,\mathcal{A},\rho)$;
\STATE $g_v\leftarrow\frac{\mathcal{L}(\theta+\varepsilon v;\xi,\mathcal{M})-\mathcal{L}(\theta-\varepsilon v;\xi,\mathcal{M})}{2\varepsilon}\,v$;
\STATE $H_k\leftarrow H_k\cup\{g_v\}$, $\mathrm{var}\leftarrow\textsc{HalfMeanVar}(H_k)$;
\STATE $z\leftarrow\textsc{CPUR}(g_v;\kappa)$;
\STATE \textbf{return} $z,\mathrm{var}$;
\end{algorithmic}
\end{algorithm}
\vspace{-6pt}

\section{Evaluation}
\label{sec:evaluation}

\subsection{Experimental Setup}
{\bf SMELL prototype.} We implement SMELL in FedML \cite{he2020fedmlresearchlibrarybenchmark} using PyTorch \cite{paszke2019pytorch} and HuggingFace Transformers \cite{wolf2020transformers}. LSFA follows the forward-gradient scheme of FwdLLM under FedAvg
\cite{FwdLLM,mcmahan2017communication}. The same FedAvg orchestration is used for the dense, sparse, and compressed configurations. The block-importance predictor follows Jenga \cite{wang2025jenga}. We pretrain it once on public data (disjoint from the test and evaluation dataset) and use it to score token blocks during local adaptation.

{\bf Model settings.} The memory study covers both LLaMA2-7B \cite{Touvron2023Llama2} and OPT-350M \cite{zhang2022optopenpretrainedtransformer}, with context lengths from 1K to 16K. LLaMA2-7B has 32 layers, a hidden dimension of 4096, 32 attention heads, and a
feed-forward dimension of 11008. OPT-350M has 24 layers, a hidden dimension of 1024, 16 attention heads, and a
feed-forward dimension of 4096. Its memory sweep uses one synthetic token sequence per context length. Both sweeps use bf16 and a batch size of one. We use LoRA rank 8 in every measured configuration that trains low-rank adapters. We use GoEmotions \cite{demszky2020goemotionsdatasetfinegrainedemotions} and Discovery \cite{sileo2019mining} for their memory runs relatively and average the measured peaks over ten runs at each length.

In a separate 16K federated setup, we also use OPT-350M to evaluate resampling and sparsity coordination. Following LongICLBench \cite{li2025longcontext}, we convert the Discovery conjunction dataset  into generative samples. Each sample
concatenates an instruction, about 260 or more demonstrations, and a query whose target is a conjunction word. We report full G-PPL, the token-weighted generative perplexity over the entire sequence. The reported score therefore reflects the model's fit to the
demonstrations and query together. A per-class Dirichlet split assigns data to 30 devices ($\alpha=0.1$ or $0.3$, seed 42). Each device holds 100 training samples, and the global test set contains 500 samples for the shared evaluation.

The classification experiments use DistilBERT-base-uncased \cite{sanh2020distilbertdistilledversionbert} at a 16K context length on the 11-class arXiv dataset \cite{ccdv2023arxivclassification}. DistilBERT has six layers, a hidden dimension of 768, 12 attention
heads, and a feed-forward dimension of 3072. We use a Dirichlet split over 30 devices with the same concentration settings ($\alpha=0.1$ or $0.3$, seed 42). This setup lets us compare how the same masking protocols behave on a classification objective. The split
operates directly on classification samples, without the demonstration concatenation used for Discovery.

{\bf Federated protocol and token sparsification.} All 30 devices participate in every federated round, and we simulate FedAvg serially. For these experiments, LoRA is applied to the query, key, value, and output projections ($r_{\mathrm{LoRA}}=1$,
$\alpha_{\mathrm{LoRA}}=2$). LSFA draws $D$ independent standard Gaussian directions across the trainable parameters and computes central differences with $\varepsilon=10^{-3}$. The server broadcasts the perturbation seeds so that all devices use the same
directions. We use either a fixed $D$ or a threshold on $\mathrm{var}$ to control sampling. Before aggregation, we clip the L2 norm of each device update at three times the median update norm to limit the effect of gradient jumps. The uplink carries update
information, block scores, and the variance statistic as required by the protocol. Raw training examples remain on the device.

On OPT-350M, we apply token sparsification in layers 11--22 and leave layers 0--10 and 23 dense. At 16K, dividing the input into blocks of 64 tokens gives 256 blocks. The keep ratio $\rho=0.4$ allocates 102 slots per sparsified layer. The local-mask baseline fills
these slots using local block scores. With CASC enabled, consensus anchors determine part of the selection, and local scores determine the remaining blocks.

{\bf Baselines.} For the memory comparison, full fine-tuning (Full FT) updates all model parameters and provides an analytical reference at lengths beyond the device budget. (i) LoRA \cite{hu2022iclr-lora} freezes the backbone and trains low-rank adapters while
retaining dense forward and backward computation. (ii) Jenga \cite{wang2025jenga} reduces activation memory through block-level token sparsification and still uses backpropagation. (iii) FwdLLM+LoRA \cite{FwdLLM,hu2022iclr-lora} estimates LoRA updates from forward passes and
serves as the BP-free memory reference. (iv) FwdLLM provides the dense reference in the federated experiments. (v) FwdLLM+Jenga applies Jenga-style local token selection at $\rho=0.4$ without consensus or uplink compression. This sparse configuration is the CASC-off
reference when we compare masking protocols at the same keep ratio.

The experiments separate device memory from the quality and communication costs of the federated updates. We first compare peak allocated memory across context lengths (Sec.~\ref{sec:eval_mem}). We then vary the number of sampled directions to assess resampling at
a matched round budget (Sec.~\ref{sec:eval_resample}). The sparsity study compares masking protocols under two non-IID partitions (Sec.~\ref{sec:eval_CASC}). The compression study examines direction fidelity and uplink cost on synthetic vectors and DistilBERT
gradients (Sec.~\ref{sec:eval_CPUR}). Finally, we evaluate the complete system over 60 rounds (Sec.~\ref{sec:e2e}) and summarize the component comparisons (Sec.~\ref{sec:ablation}).

\subsection{Memory Reduction Performance}\label{sec:eval_mem}
{\bf Metric and setup.} On LLaMA2-7B, we compare SMELL with Full FT, LoRA, Jenga, and FwdLLM+LoRA. On OPT-350M, Table~\ref{tab:mem_peak} includes LoRA and FwdLLM+LoRA, and we quantify SMELL's savings relative to FwdLLM+LoRA. Each measured run uses one input at the
specified context length. We report the peak memory allocated to tensors during the training computation. This measure captures the largest allocation at any point during a run and excludes unused memory held in reserve by the allocator for later allocations.

{\bf Results and conclusion.} Table~\ref{tab:mem_peak} reports peak memory in GB. Each Scale entry divides the method's memory by SMELL's memory at the same context length. For LLaMA2-7B, the Full FT and LoRA entries use the analytical estimate in
Sec.~\ref{sec:mem_consumption}. The other LLaMA2-7B entries come from measurements, with out-of-memory runs marked against the 40~GB usable budget. The analytical entries provide a reference for configurations beyond the device budget. We interpret each ratio
within a model and at a fixed context length so that differences in model size do not drive the comparison.

At 16K, SMELL uses 27.5~GB on LLaMA2-7B, a reduction of 88.35\% relative to estimated Full FT and 13.25\% relative to measured FwdLLM+LoRA. Jenga fits the budget through 4K and runs out of memory at 8K and 16K. On OPT-350M, SMELL uses less memory than FwdLLM+LoRA
at every tested length. The relative saving reaches 16.92\% at 16K. For both models, the savings relative to FwdLLM+LoRA increase across the measured context lengths. The growing gap indicates that sparse forward computation becomes more valuable as the input grows
within this sweep.

\begin{table*}[t]
\centering
\renewcommand{\arraystretch}{1.15}
\caption{Peak allocated memory in GB for LLaMA2-7B and OPT-350M in bf16, with LoRA rank $r=8$ where applicable.}
\vspace{-0.1in}
\label{tab:mem_peak}
\setlength{\tabcolsep}{4.8pt}
\begin{tabular}{c|cc|cc|cc|cc|c>{\columncolor{lb}[0pt][0pt]}c|cc|cc|c>{\columncolor{lb}[0pt][0pt]}c}
\hline
\multicolumn{1}{c|}{} & \multicolumn{10}{c|}{LLaMA2-7B} & \multicolumn{6}{c}{OPT-350M}\\
\hline
Context & \multicolumn{2}{c|}{Full FT} & \multicolumn{2}{c|}{LoRA} & \multicolumn{2}{c|}{Jenga} & \multicolumn{2}{c|}{FwdLLM+LoRA} & \multicolumn{2}{c|}{SMELL (Ours)} & \multicolumn{2}{c|}{LoRA} & \multicolumn{2}{c|}{FwdLLM+LoRA} & \multicolumn{2}{c}{SMELL (Ours)}\\
 & Scale & Mem & Scale & Mem & Scale & Mem & Scale & Mem & Scale & \colorbox{white}{Mem} & Scale & Mem & Scale & Mem & Scale & \colorbox{white}{Mem}\\
\hline
1K  & 8.19$\times$ & 113 & 1.52$\times$ & 21 & 1.19$\times$ & 16.5 & 1.01$\times$ & 13.9 & 1$\times$ & \textbf{13.8} & 2.01$\times$ & 2.00 & 1.03$\times$ & 1.02 & 1$\times$ & \textbf{0.99}\\
2K  & 8.28$\times$ & 121 & 2.05$\times$ & 30 & 1.38$\times$ & 20.2 & 1.02$\times$ & 15.0 & 1$\times$ & \textbf{14.6} & 2.60$\times$ & 3.28 & 1.04$\times$ & 1.31 & 1$\times$ & \textbf{1.26}\\
4K  & 8.20$\times$ & 137 & 2.75$\times$ & 46 & 1.66$\times$ & 27.8 & 1.03$\times$ & 17.2 & 1$\times$ & \textbf{16.7} & 3.51$\times$ & 4.89 & 1.08$\times$ & 1.50 & 1$\times$ & \textbf{1.39}\\
8K  & 8.25$\times$ & 170 & 3.83$\times$ & 79 & --- & $>$40 & 1.08$\times$ & 22.2 & 1$\times$ & \textbf{20.6} & 4.86$\times$ & 8.10 & 1.14$\times$ & 1.89 & 1$\times$ & \textbf{1.66}\\
16K & 8.58$\times$ & 236 & 5.27$\times$ & 145 & --- & $>$40 & 1.15$\times$ & 31.7 & 1$\times$ & \textbf{27.5} & 6.56$\times$ & 14.51 & 1.20$\times$ & 2.66 & 1$\times$ & \textbf{2.21}\\
\hline
\end{tabular}

\vspace{2pt}
{\footnotesize Scale is relative to SMELL at the same length. \colorbox{lb}{\textbf{Blue}} marks the lowest memory for each model and context length.}
\vspace{-0.2in}
\end{table*}

\subsection{Gradient Resampling Performance}\label{sec:eval_resample}
{\bf Metric and setup.} We vary the number of sampled directions on OPT-350M at 16K to measure the effect of resampling on full G-PPL. Averaging independent estimates gives the $1/D$ variance scaling described by LSFA. Here we fix $D\in\{16,22,32\}$ and update the
trainable parameters across all layers in each round. The variance threshold is used in the end-to-end experiment in Sec.~\ref{sec:e2e}. Each setting uses the same training and evaluation protocol. A matched round budget isolates the convergence gain per round,
while a larger $D$ still requires more forward evaluations.

{\bf Results and conclusion.} Table~\ref{tab:resample} compares the first eight rounds using the same pre-r0 full G-PPL as a shared baseline. By the end of round 7 (i.e., r7), both $D\!=\!22$ and $D\!=\!32$ achieve greater improvements than $D\!=\!16$. The two larger budgets yield similar endpoints over this initial window, so the early comparison mainly separates them from the smallest budget. Over 30 rounds, $D\!=\!32$ reaches 31.410, a 6.70\% reduction from the shared baseline, while $D\!=\!16$ reaches 32.260, a 4.18\% reduction. The longer comparison shows that the gain from more sampled directions persists beyond the first eight rounds at the same round budget.

\begin{table}[t]
\centering
\renewcommand{\arraystretch}{1.15}
\caption{Resampling directions $D$ over a unified 8-round window.}
\label{tab:resample}
\vspace{-0.1in}
\setlength{\tabcolsep}{8pt}
\begin{tabular}{c|c|c|c|c|c}
\hline
$D$ & Window & Pre-r0 & End & $\Delta$ & $\Delta$\%\\
\hline
16 & pre-r0--r7 & 33.6653 & 32.7144 & $-0.9510$ & $-2.82$\\
22 & pre-r0--r7 & 33.6653 & 32.5335 & $-1.1318$ & $-3.36$\\
\rowcolor{lb} 32 & pre-r0--r7 & 33.6653 & \textbf{32.5071} & \textbf{$-1.1582$} & \textbf{$-3.44$}\\
\hline
\end{tabular}

\vspace{2pt}

\scriptsize \colorbox{lb}{\textbf{Blue}} marks the lowest full G-PPL at r7 among the three direction budgets.
\end{table}

\subsection{Sparsity Coordination Performance}
\label{sec:eval_CASC}
{\bf Metric and setup.} The server sums the block scores from participating devices and uses the top and bottom fractions ($r=0.2$) to set positive and negative consensus anchors. The 51 positive anchors occupy half of the 102 retained slots. The 51 negative
anchors are excluded, and each device fills the remaining 51 slots using its local scores. We compare three protocols at the same sparsity budget: CASC off (i.e. $\mathcal{A}_i$ set to $0$ for every block $i$), CASC on during training only, and CASC on during both training and evaluation.

On DistilBERT, we report negative log-likelihood (NLL) as the primary metric and accuracy as a secondary metric. Let $N$ be the number of test samples, $o_m(\xi)$ the logit of class $m$ for sample $\xi$, and $\mathcal{Y}$ the set of 11 labels. We compute
$\mathrm{NLL}:=-\frac{1}{N}\sum_{(\xi,y)\in\mathcal{D}_{\text{test}}}\log\frac{\exp(o_y(\xi))}{\sum_{m\in\mathcal{Y}}\exp(o_m(\xi))}$. NLL evaluates the predicted probability assigned to the true label, while accuracy counts correct top predictions.

\begin{table}[t]
\centering
\renewcommand{\arraystretch}{1.10}
\caption{CASC ablations after 30 rounds on (a) OPT-350M/Discovery and (b) DistilBERT/arXiv, each evaluated on 500 test samples.}
\vspace{-0.1in}
\label{tab:main}
\footnotesize
\setlength{\tabcolsep}{3pt}
\begin{tabular}{c|l|c|c|c|c}
\hline
\multicolumn{6}{c}{\textit{(a) OPT-350M/Discovery (full G-PPL)}}\\
\hline
$\alpha$ & Protocol & r0(G-PPL) & r29(G-PPL) & $\Delta$ & $\Delta$\%\\
\hline
0.1 & CASC off & 33.525 & 31.533 & $-1.992$ & $-5.94$\\
\rowcolor{lb} 0.1 & CASC on(train) & 33.525$^{*}$ & \textbf{31.378} & \textbf{$-2.147$} & \textbf{$-6.41$}\\
0.1 & CASC on(train+eval) & 33.921 & 31.635 & $-2.286$ & $-6.74$\\
0.3 & CASC off & 33.530 & 31.457 & $-2.073$ & $-6.18$\\
\rowcolor{lb} 0.3 & CASC on(train) & 33.530$^{*}$ & \textbf{31.416} & \textbf{$-2.113$} & \textbf{$-6.30$}\\
0.3 & CASC on(train+eval) & 33.940 & 31.696 & $-2.244$ & $-6.61$\\
\hline
\hline
\multicolumn{6}{c}{\textit{(b) DistilBERT/arXiv (NLL and accuracy)}}\\
\hline
$\alpha$ & Protocol & r0(NLL) & r29(NLL) & $\Delta$ & Acc\\
\hline
0.1 & CASC off & 1.6910 & 1.5836 & $-0.1075$ & 0.468\\
0.1 & CASC on(train) & 1.6910$^{*}$ & 1.5768 & $-0.1142$ & 0.460\\
\rowcolor{lb} 0.1 & CASC on(train+eval) & 1.7527 & \textbf{1.5563} & $-0.1964$ & 0.458\\
0.3 & CASC off & 1.6906 & 1.5887 & $-0.1019$ & 0.466\\
0.3 & CASC on(train) & 1.6906$^{*}$ & 1.5772 & $-0.1134$ & 0.462\\
\rowcolor{lb} 0.3 & CASC on(train+eval) & 1.7511 & \textbf{1.5484} & $-0.2026$ & 0.454\\
\hline
\end{tabular}

\vspace{2pt}
{\scriptsize $^{*}$Shared unmasked r0 anchor. CASC off uses local masks. Train and train+eval apply consensus in training only or in training and evaluation, respectively. \colorbox{lb}{\textbf{Blue}} marks the lowest r29 value of the primary metric within each partition.}
\end{table}

{\bf Results and conclusion.} Table~\ref{tab:main} reports endpoint r29 (i.e. the metrics after 30 rounds). Each $\Delta$ uses the r0 evaluation anchor for its own row. On OPT-350M, enabling CASC during training lowers r29 full G-PPL relative to CASC off at both $\alpha$ values when evaluation uses the same unmasked readout. Applying the consensus mask during evaluation changes the r0 anchor and
leaves a higher absolute r29 full G-PPL than the local-mask baseline, despite a larger reduction from its own r0. The matched evaluation protocol is therefore essential when interpreting the OPT-350M endpoints.

On DistilBERT, both CASC protocols lower NLL at r29 for both non-IID partitions, with the lowest NLL obtained when the consensus mask is also used at evaluation. Accuracy is slightly lower than with local masks. The two metrics therefore show a trade-off: the CASC
arms improve the mean log-probability of the true label while producing slightly fewer correct top predictions than the local-mask baseline.

\begin{figure*}[t]
    \centering
    \begin{minipage}[t]{0.242\linewidth}
        \centering
        \includegraphics[width=\linewidth]{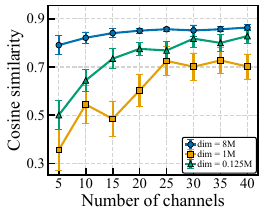}
        \par\vspace{-6pt}
        {\scriptsize (a) Compression rate $=1/8$}
    \end{minipage}\hfill
    \begin{minipage}[t]{0.242\linewidth}
        \centering
        \includegraphics[width=\linewidth]{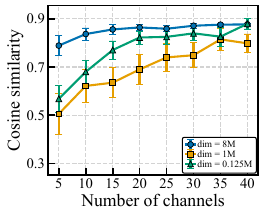}
        \par\vspace{-6pt}
        {\scriptsize (b) Compression rate $=1/4$}
    \end{minipage}\hfill
    \begin{minipage}[t]{0.242\linewidth}
        \centering
        \includegraphics[width=\linewidth]{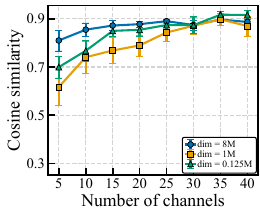}
        \par\vspace{-6pt}
        {\scriptsize (c) Compression rate $=1/2$}
    \end{minipage}\hfill
    \begin{minipage}[t]{0.242\linewidth}
        \centering
        \includegraphics[width=\linewidth]{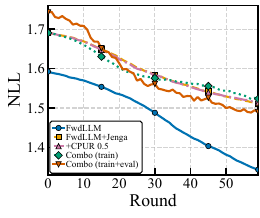}
        \par\vspace{-6pt}
        {\scriptsize (d) DistilBERT end-to-end NLL}
    \end{minipage}
    \caption{(a)--(c) Cosine similarity versus the number of accumulated channels for three vector dimensions and per-channel rates of $1/8$, $1/4$, and $1/2$. Error bars show 95\% $t$-based confidence intervals over 50 trials. (d) End-to-end NLL on DistilBERT/arXiv ($\alpha=0.3$) over 60 rounds. Combo (train) and Combo (train+eval) share one training run. The two curves differ in whether the consensus mask is applied when scoring the test set.}
    \label{fig:vq_cos}
    \label{fig:e2e}
    \vspace{-0.05in}
\end{figure*}
\subsection{Compressed Transmission Performance}
\label{sec:eval_CPUR}
{\bf Metric and setup.} We evaluate CPUR's Count-Sketch representation on synthetic vectors and on gradients from DistilBERT trained on arXiv. For an input dimension $d$, each channel contains $d_c=\mathrm{rate}\times d$ buckets, and $c$ denotes the number of
channels available for recovery. Cosine similarity between the original and reconstructed vectors measures the fidelity of the update direction. We also time a complete compression and decompression pass on a single CPU thread. Timing a full pass and measuring
uplink payload answer different questions. The first characterizes the implementation cost, while the second measures the data that each device sends in each compressed transmission to the server.

{\bf Results and conclusion.} We draw synthetic vectors from a standard normal distribution and average each point over 50 trials. Fig.~\ref{fig:vq_cos}(a)--(c) shows that accumulating more channels generally improves cosine similarity, with diminishing gains near
the plateau. Increasing the per-channel rate also generally improves fidelity. For the largest dimension, the curves at different rates can cross once similarity has leveled off. For $d=1$M and $d=0.125$M, the higher rates consistently yield higher similarity
across the displayed range.

On real gradients, we use per-result Count-Sketch at rate 0.5 in the DistilBERT LoRA experiment (bf16, 64 channels, one per direction--sample pair in the $8\times8$ setting). Over 60 rounds, CPUR alone halves total uplink traffic relative to the uncompressed
local-mask baseline. Table~\ref{tab:e2e} shows an NLL increase of 0.0029 and unchanged endpoint accuracy. The aggregated cosine similarity is 0.9945. In this configuration, the large reduction in payload comes with a small change in endpoint NLL relative to the
sparse reference.

These results support progressive recovery across LSFA's resampling rounds. Each transmission carries one Count-Sketch channel, and the server averages the recovered estimates from $c$ independently hashed channels as in Eq.~\ref{eq:decompress}. A smaller
per-channel rate reduces the size of each transmission. Accumulating multiple channels improves the recovered direction in the tested settings, with the synthetic sweep showing where the gains begin to level off as channels accumulate.

\subsection{End-to-End System Performance}
\vspace{-0.03in}
\label{sec:e2e}
{\bf Metric and setup.} We run the full system on DistilBERT with 30 devices for 60 rounds. LSFA uses $\rho=0.4$ and a $\mathrm{var}$ threshold of 0.587, CASC uses $r=0.2$, and CPUR uses $\mathrm{rate}=0.5$. We choose the threshold so that the average number of
resampling passes is approximately matched to the baseline. Matching the average sampling effort helps separate the combined effect of consensus and compression from the effect of drawing more directions per round. We  compare dense FwdLLM, the
uncompressed local-mask baseline, the CPUR-only variant, and the full system. The two full-system rows share one training run and differ only in the use of the consensus mask when scoring the same held-out examples in the held-out arXiv classification set.

{\bf Results and conclusion.} As shown in Table~\ref{tab:e2e} and Fig.~\ref{fig:e2e}(d), with the consensus mask enabled during evaluation, the full SMELL system achieves an NLL of 1.4896 and an accuracy of 0.478 at round 59, outperforming FwdLLM+Jenga (NLL: 1.5110; accuracy: 0.474) while reducing uplink traffic by 50\%. However, disabling the mask during evaluation increases the NLL to 1.5231, exceeding that of the sparse baseline, indicating that the performance gain depends on the evaluation protocol. Since both full-system configurations share the same trained model, their performance difference isolates the effect of evaluation-time masking. Nevertheless, dense FwdLLM achieves a lower NLL of 1.3437, highlighting the remaining accuracy cost of sparse fine-tuning. Overall, these results demonstrate the potential of SMELL to improve communication efficiency while maintaining competitive performance within the evaluated sparse fine-tuning setting.

\begin{table}[!htbp]
\centering
\renewcommand{\arraystretch}{1.15}
\caption{End-to-end evaluation on DistilBERT/arXiv with $\alpha=0.3$. We report r59 results after 60 rounds on 500 test samples.}
\label{tab:e2e}
\small
\setlength{\tabcolsep}{2.5pt}
\begin{tabular}{l|c|c|c|c|c}
\hline
Run & NLL r0 & NLL r59 & $\Delta$ & Acc r59 & Uplink\\
\hline
\rowcolor{pink}FwdLLM & 1.5923 & 1.3437 & $-0.2486$ & 0.520 & 1.0$\times$\\
FwdLLM+Jenga & 1.6906 & 1.5110 & $-0.1797$ & 0.474 & 1.0$\times$\\
+ CPUR 0.5 & 1.6907 & 1.5139 & $-0.1768$ & 0.474 & \textbf{0.50$\times$}\\
Combo (train) & 1.6904 & 1.5231 & $-0.1673$ & 0.470 & \textbf{0.50$\times$}\\
\rowcolor{lb}Combo (train+eval) & 1.7425 & {\bf 1.4896} & \textbf{$-0.2529$} & {\bf 0.478} & \textbf{0.50$\times$}\\
\hline
\end{tabular}

\vspace{2pt}
{\footnotesize Baseline = CASC off (local mask, no consensus); CPUR = progressive Count-Sketch compression; Combo = variance-aware resampling + CASC + CPUR; train = mask in training only; train+eval = mask in training and evaluation; $\Delta$ is r0$\to$r59; train+eval uses its own masked r0 (1.7425). \colorbox{lb}{\textbf{Blue}} means the best in-family run. \colorbox{pink}{\textbf{Red}} means the naive FwdLLM baseline.}
\end{table}
\vspace{-0.05in}
\subsection{Ablation Study and Conclusion}
\label{sec:ablation}
\vspace{-0.03in}
{\bf Metric and setup.} We use the earlier comparisons to assess the three mechanisms under the shared 16K protocol. The CASC study holds the keep ratio fixed and varies when consensus anchors are applied (Sec.~\ref{sec:eval_CASC}). The CPUR-only arm adds
compression to the local-mask baseline, providing a direct comparison of uplink and endpoint quality (Sec.~\ref{sec:eval_CPUR}). The dense and sparse arms in Sec.~\ref{sec:e2e} show how local token sparsification changes the endpoint before consensus and
compression are combined in the full pipeline.

{\bf Results and conclusion.} CASC improves the primary performance metric on both models when sparsity coordination is applied during training but disabled during evaluation, demonstrating the benefits of incorporating cross-device consensus into local token selection. However, applying the consensus mask during evaluation has different effects on the full G-PPL of OPT-350M and the NLL of DistilBERT (Table~\ref{tab:main}). These results highlight the importance of distinguishing training-time and evaluation-time masking when interpreting performance differences across configurations. Regarding communication efficiency, CPUR substantially reduces uplink traffic with only a marginal increase in NLL and no loss in classification accuracy in the DistilBERT experiment (Table~\ref{tab:e2e}). Furthermore, under the masked-evaluation protocol, the complete SMELL framework outperforms the uncompressed sparse reference while retaining the communication savings, demonstrating the complementary benefits of sparsity coordination and compressed update recovery. Nevertheless, the dense FwdLLM baseline maintains superior performance on the classification task, reflecting the accuracy cost associated with aggressive token sparsification. Notably, most of the performance gap is already observed in the uncompressed sparse configuration, whereas CPUR alone introduces only a limited additional degradation. This distinction suggests that token sparsification, rather than update compression, is the primary source of the observed accuracy–efficiency trade-off. Meanwhile, CASC helps mitigate the performance loss through cross-device sparsity coordination, and CPUR reduces communication demands while largely preserving the quality of sparse model updates. These findings clarify the individual contributions of the proposed mechanisms and demonstrate how SMELL integrates memory-efficient sparse computation, heterogeneity-aware coordination, and compressed update transmission within a unified federated fine-tuning framework.

\section{Related Work}
\label{sec:related_work}

{\bf Federated fine-tuning and PEFT.} Federated learning \cite{mcmahan2017communication} enables collaborative model adaptation without centralizing raw training data, and recent work \cite{wu2026survey} has systematized its integration with LLM fine-tuning. LoRA \cite{hu2022iclr-lora} reduces trainable parameters and update payloads, but heterogeneous data and device resources complicate federated adaptation. FedALT \cite{bian2026fedalt} combines a client's individual LoRA with a separate adapter representing other clients, using an adaptive mixer to balance personalization and shared knowledge. Fed-PLoRA \cite{zhang2026fedplora} uses parallel rank-one modules and a Select-N-Fold strategy to support clients with different resource budgets while reducing initialization and aggregation noise. These methods coordinate parameter adaptation across clients; SMELL instead focuses on coordinating the contextual token selections used to construct sparse local updates. Sketch-based methods such as FetchSGD \cite{rothchild2020fetchsgdcommunicationefficientfederatedlearning} reduce communication by aggregating compressed client updates, whereas CPUR couples compressed channels with repeated forward-gradient estimation for progressive update recovery.

{\bf Long-context and memory-efficient fine-tuning.} LongLoRA \cite{chen2023longlora} reduces the cost of long-context adaptation through shifted sparse attention. Recent work further targets the activations retained during adaptation: LARS \cite{tenison2026lars} performs adaptation in sequence-compressed low-rank subspaces to reduce stored adapter intermediates. Seeger et al. \cite{seeger2026forget} fine-tune models under sparse KV-cache policies so that model parameters can adapt to the chosen attention policy. Adjacent efficiency approaches operate at different levels: SeCO/SpaCO \cite{li2025chunkwise} reorganize chunk-wise backpropagation and selectively propagate gradients through chunks, while NSA \cite{yuan2025nsa} co-designs a natively trainable sparse-attention mechanism with hardware-efficient execution. These approaches differ from the contextual token elimination used in SMELL. Most directly related is Jenga \cite{wang2025jenga}, which predicts token-block importance and exploits contextual token sparsity to reduce long-context activation memory. SMELL builds on this principle while combining sparse forward computation with BP-free adaptation and cross-device consensus anchors. Our preliminary experiments examine the memory peak of the evaluated Jenga configurations and the divergence of locally selected masks under non-IID data (Sec.~\ref{sec:motivation_non_iid}), motivating coordination between shared token preferences and local selection flexibility.

{\bf BP-free and zeroth-order optimization.} Forward-mode differentiation \cite{baydin2022gradientsbackpropagation} computes directional derivatives without reverse-mode backpropagation, while MeZO \cite{malladi2023mezo} uses finite-difference estimates for memory-efficient LLM fine-tuning. MUZO \cite{peng2025muzo} combines multiple queries with momentum-based optimization to improve zeroth-order estimation and convergence. In the field of federated learning, FwdLLM \cite{FwdLLM} uses multiple perturbed inferences to reduce client memory. More recent methods address several resource and optimization constraints together: FedSPZO \cite{ahmed2026fedspzo} reuses intermediate activations across block-wise perturbations to reduce repeated forward computation; ZorBA \cite{meng2026zorba} assigns heterogeneous subsets of Transformer blocks to clients and uses shared seeds for compact communication; and Chen et al. \cite{chen2026global} combine local personalized perturbations with global adaptive momentum to handle heterogeneous data. FeedSign \cite{cai2026feedsign} further reduces communication through shared random generators and sign-based updates, while HeimdaLLM \cite{HeimdaLLM2026} uses BP-based cloud-assisted gradient rectification to improve zeroth-order estimates on edge devices. SMELL addresses a different coordination problem: CASC coordinates input-dependent token-block masks, and CPUR recovers updates from repeated sparse-forward estimates. The coordination concerns contextual selections rather than the assignment of trainable Transformer blocks or personalized perturbation distributions.

SMELL brings these directions together around federated contextual token sparsity: LSFA provides sparse forward-only adaptation, CASC balances shared token preferences with local flexibility, and CPUR supports compressed update recovery. This design targets the interaction between sparse computation, heterogeneous local data, and communication constraints in long-context federated fine-tuning.

\section{Conclusion}
\label{sec:conclusion}
\vspace{-0.05in}

We presented SMELL, a sparsity-aware and memory-efficient framework integrating LSFA, CASC, and CPUR for federated long-context LLM fine-tuning on resource-constrained devices. SMELL addresses memory demands, data heterogeneity, and communication overhead through sparse-forward adaptation, cross-device sparsity coordination, and compressed update recovery. Experimental results demonstrate that LSFA substantially reduces memory consumption while maintaining effective model adaptation, CASC improves fine-tuning performance under non-IID settings, and CPUR enables lightweight transmission with progressive update reconstruction. These results highlight the value of jointly optimizing sparse computation, federated coordination, and update compression for resource-efficient long-context adaptation.

\section*{AI-Generated Content Acknowledgement}
The authors used the DeepSeek large language model, accessed through the OpenCode agent, to assist with paper polishing and the development of the experimental code. All experimental data and results come from the authors' own experiments, and the authors reviewed and verified all scientific content and take full responsibility for the content of the publication.

\bibliographystyle{IEEEtran}
\bibliography{refs}

\end{document}